\documentclass{article}
\usepackage{ijcai26}

\usepackage{CJKutf8} 
\usepackage{xcolor}
\usepackage{graphicx}
\usepackage{booktabs}

\usepackage{times}
\usepackage{soul}
\usepackage{url}
\usepackage[hidelinks]{hyperref}
\usepackage[utf8]{inputenc}
\usepackage[small,labelfont=normalfont,textfont=normalfont]{caption}
\usepackage{graphicx}
\usepackage{amsmath}
\usepackage{amsthm}
\usepackage{booktabs}
\usepackage{algorithm}
\usepackage{algorithmic}
\usepackage[switch]{lineno}

\usepackage{amsmath, amssymb}
\usepackage{algorithm}
\usepackage{xcolor} % 为了给代码添加颜色，可选
\usepackage{listings}
\usepackage{algorithmic}

\usepackage{booktabs} % For professional-looking tables
\usepackage{siunitx}  % For aligning numbers by decimal point

\usepackage{tikz}
\usetikzlibrary{shapes, arrows.meta, positioning, backgrounds, fit, calc, shadows}
\usepackage{amsmath}
\usepackage{amssymb}

\title{G$^{2}$C-MT: Graph-Guided Context Selection for Document-Level Machine Translation}

\author{
Baijun Ji
\and
Zixuan Zhou
\and
Xiangyu Duan
\and
Yu Liu
\and
Longbo Sun
\and
Rupu Wei
\And
Bohong Zhao
\affiliations
School of Computer Science and Technology, Soochow University\\
Trip.com Group
\emails
cocaer.cl@gmail.com, zxzhou1213@stu.suda.edu.cn, xyduan@suda.edu.cn,\\
\{liu.yub, lbsun, rpwei, bohongzhao\}@trip.com
}

\begin{document}

\maketitle
\begin{abstract}
Effective document-level machine translation (DocMT) requires capturing long-range discourse dependencies. Recent work has explored retrieval-based and discourse-aware context selection. However, these approaches often lack an explicit mechanism for modeling structured discourse dependencies between distant paragraphs in a document.
In this paper, we propose G²C-MT (Graph-Guided Context for Machine Translation),
which views DocMT context selection as a structured path discovery problem on a lightweight discourse graph,
rather than retrieving unstructured context sets or relying on expensive LLM-based discourse modeling.
In detail, we represent each paragraph as a node and model the relationship between each pair of nodes, considering their semantic similarity, adjacency, and keyword overlap.
Furthermore, we propose a depth-biased random walk over the graph to sample a backward context path for each target paragraph. The context path will be used to prompt a large language model (LLM) for translation.
This framework naturally supports multi-path context sampling, which can improve robustness by aggregating diverse translation candidates for discourse-ambiguous inputs.
Experiments conducted across various domains show that G²C-MT outperforms strong baselines on multiple LLMs, including DeepSeek-V3, Gemini-2.5-Flash-lite, and the Qwen-2.5/3 series.
\end{abstract}

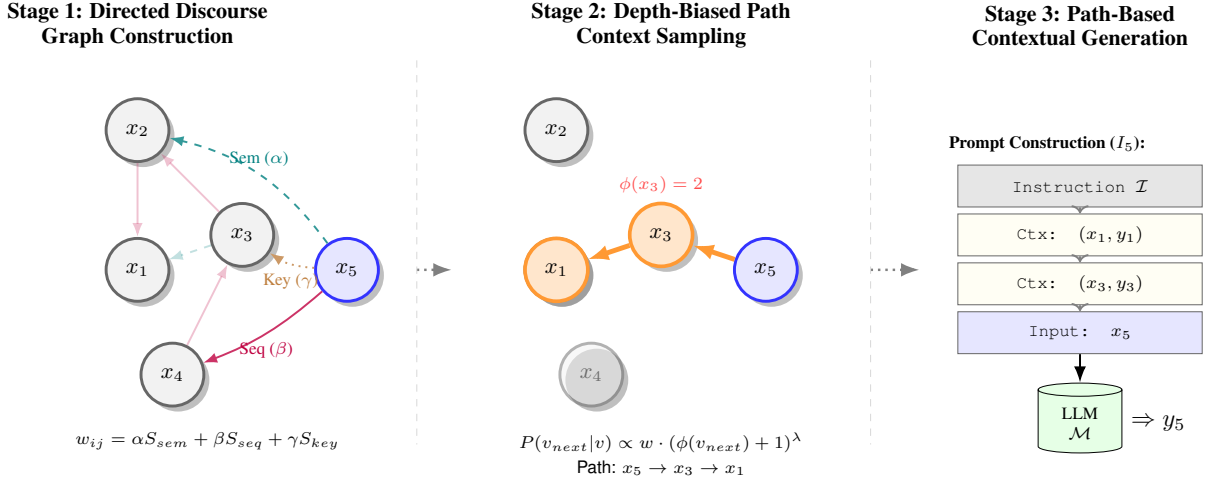
\begin{figure*}[htbp]
\centering
\resizebox{0.9\textwidth}{!}{% 缩放以适应页面宽度
\begin{tikzpicture}[
    font=\sffamily\small,
    node distance=1.2cm and 1.5cm,
    paragraph/.style={circle, draw=black!60, very thick, minimum size=0.9cm, fill=white, drop shadow},
    target/.style={paragraph, fill=blue!10, draw=blue!80},
    context/.style={paragraph, fill=gray!10},
    selected/.style={paragraph, fill=orange!20, draw=orange!80, line width=1.5pt},
    % 边样式
    edge_sem/.style={-{Latex[length=2mm]}, draw=teal!80, dashed, thick},
    edge_seq/.style={-{Latex[length=2mm]}, draw=purple!80, thick},
    edge_key/.style={-{Latex[length=2mm]}, draw=brown!80, dotted, thick},
    path_edge/.style={-{Latex[length=3mm]}, draw=orange!80, line width=2pt},
    % 文本框样式
    box_style/.style={rectangle, draw=black!40, rounded corners, inner sep=6pt, align=center, fill=white, drop shadow},
    prompt_box/.style={rectangle, draw=black!60, fill=yellow!5, inner sep=4pt, minimum height=0.6cm, font=\scriptsize\ttfamily},
    label_style/.style={font=\bfseries\small, align=center}
]

    % ==========================================================
    % STAGE 1: Look-Back Discourse Graph
    % ==========================================================
    
    % 背景框
    \node[label_style] (stage1_label) at (0, 3.5) {Stage 1: Directed Discourse \\ Graph Construction};
    
    % 节点布局 (模拟流形结构)
    \node[context] (x1) at (0, 0) {$x_1$};
    \node[context] (x2) at (0, 2) {$x_2$};
    \node[context] (x3) at (1.5, 0.5) {$x_3$};
    \node[context] (x4) at (0.5, -1.5) {$x_4$};
    \node[target] (x5) at (3, 0) {$x_5$}; % Target node
    
    % 边 (展示三种连接)
    \draw[edge_seq] (x5) to[bend left=10] node[midway, below, font=\scriptsize, text=purple] {Seq ($\beta$)} (x4); % 顺序
    \draw[edge_sem] (x5) to[bend right=20] node[midway, above, font=\scriptsize, text=teal] {Sem ($\alpha$)} (x2); % 语义
    \draw[edge_key] (x5) to[bend left=15] node[midway, below, font=\scriptsize, text=brown] {Key ($\gamma$)} (x3); % 词汇
    \draw[edge_sem, opacity=0.3] (x3) -- (x1);
    \draw[edge_seq, opacity=0.3] (x2) -- (x1);
    \draw[edge_seq, opacity=0.3] (x4) -- (x3);
    \draw[edge_seq, opacity=0.3] (x3) -- (x2);
    % 说明公式
    \node[anchor=north] at (1, -2.2) {\scriptsize $w_{ij} = \alpha S_{sem} + \beta S_{seq} + \gamma S_{key}$};

    % ==========================================================
    % STAGE 2: Potential-Guided Random Walk
    % ==========================================================
    
    % 移动坐标系
    \begin{scope}[xshift=6cm]
        \node[label_style] (stage2_label) at (1.5, 3.5) {Stage 2: Depth-Biased Path\\Context Sampling };
        
        % 复制节点位置
        \node[context] (p1) at (0, 0) {$x_1$};
        \node[context, label={[red!70, font=\scriptsize]}] (p2) at (0, 2) {$x_2$};
        \node[selected, label={[red!70, font=\scriptsize]above:$\phi(x_3)=2$}] (p3) at (1.5, 0.5) {$x_3$};
        \node[context, opacity=0.5] (p4) at (0.5, -1.5) {$x_4$};
        \node[target] (p5) at (3, 0) {$x_5$}; 
        
        % 路径高亮
        \draw[path_edge] (p5) -- (p3);
        \draw[path_edge] (p3) -- (p1);
        \node[selected, label={[red!70, font=\scriptsize]}] at (0, 0) {$x_1$}; % 重绘被覆盖的节点
        
        % 游走公式标注
        \node[anchor=north, align=center] at (1.5, -2.2) {\scriptsize $P(v_{next}|v) \propto w \cdot (\phi(v_{next})+1)^\lambda$ \\ \scriptsize Path: $x_5 \rightarrow x_3 \rightarrow x_1$};
        
        % 箭头表示过渡
        \draw[-{Latex}, gray, dotted, line width=1pt] (-2, 0) -- (-1.5, 0);
    \end{scope}

    % ==========================================================
    % STAGE 3: Graph-Informed Translation
    % ==========================================================
    
    \begin{scope}[xshift=12cm]
        \node[label_style] (stage3_label) at (1.5, 3.5) {Stage 3: Path-Based \\Contextual Generation};
        
        % 提示词构建 (Stack)
        \node[anchor=west, font=\scriptsize\bfseries] at (-0.5, 1.8) {Prompt Construction ($I_5$):};
        
        \node[prompt_box, fill=gray!20, minimum width=3.5cm] (inst) at (1.5, 1.2) {Instruction $\mathcal{I}$};
        \node[prompt_box, minimum width=3.5cm] (ctx1) at (1.5, 0.5) {Ctx: $(x_1, y_1)$};
        \node[prompt_box, minimum width=3.5cm] (ctx3) at (1.5, -0.2) {Ctx: $(x_3, y_3)$};
        \node[prompt_box, fill=blue!10, minimum width=3.5cm] (curr) at (1.5, -0.9) {Input: $x_5$};
        
        % 连线表示序列化
        \draw[->, thick, gray] (inst) -- (ctx1);
        \draw[->, thick, gray] (ctx1) -- (ctx3);
        \draw[->, thick, gray] (ctx3) -- (curr);
        
        % LLM 图标
        \node[draw, cylinder, shape border rotate=90, aspect=0.25, fill=green!10, minimum height=1cm, minimum width=1.2cm, align=center, font=\scriptsize] (llm) at (1.5, -2.2) {LLM\\ $\mathcal{M}$};
        
        \draw[-{Latex}, thick] (curr) -- (llm);
        \node[right, font=\bfseries] at (llm.east) {$\Rightarrow y_5$};
        
        % 箭头表示过渡
        \draw[-{Latex}, gray, dotted, line width=1pt] (-1.5, 0) -- (-0.8, 0);
    \end{scope}

    % 分隔线 (可选)
    \draw[gray!30, dashed] (4, 3) -- (4, -2.5);
    \draw[gray!30, dashed] (10.5, 3) -- (10.5, -2.5);

\end{tikzpicture}
}
\caption{\textbf{Overview of the G$^2$C-MT Framework.} The process involves three stages: (1) Constructing a discourse graph considering semantic ($\alpha$), sequential ($\beta$), and keyword ($\gamma$) cohesion; 
(2) Traversing a context path via Depth-Biased Random Walk (backtracking from target to history), where nodes with higher depth potential $\phi$ (e.g., $x_3$) and edge score attract the walker; 
(3) Formatting the path into a structured prompt for translation.}
\label{fig:overview}
\end{figure*}

\section{Introduction}
High-quality document-level machine translation requires more than accurate sentence-level translation.
It also needs the preservation of discourse phenomena, including lexical consistency and coreference resolution.
Capturing long-range dependencies is therefore essential.
Although recent advances in LLMs have shown success in handling long contexts \cite{liu2023lostmiddlelanguagemodels,chen2023longlora,gao2025train}, translating an entire document in a single pass often leads to issues such as sentence omissions or context dilution \cite{wang2025delta}. Moreover, exposing the model to the whole document context is inefficient, since the decoding cost of LLMs increases quadratically with the length of the input text.
To address this issue, recent studies have employed retrieval-based and graph-based strategies to select prior translated paragraphs as context. 
Retrieval-based methods select historical translations according to semantic similarity to mitigate long-range dependencies \cite{wang2025delta}. 
However, these methods often fail to preserve explicit discourse structures, as they treat sentences as an unstructured collection.
Similarly, existing graph-based methods depend on expensive edge relation definitions via LLM-based relation classification,
and they are usually limited to selecting first-order neighbors as historical context \cite{dutta2025graft,pham2025discourse}. 
This restriction prevents them from capturing deep and multi-hop discourse paths that define the global document structure.

To overcome the above problem, we propose G$^{2}$C-MT, a novel Graph-Guided Context for Machine Translation framework. Unlike retrieving sentences in isolation or relying   on expensive graph construction processes, we model the document’s discourse structure as a weighted directed acyclic graph (DAG) using a lightweight graph construction procedure. Specifically, each paragraph serves as a node, while edges denote the relationship between paragraphs. These relationships are quantified through a fusion score that derived from semantic similarity, sequential adjacency, and lexical overlap.
This rich metric enables our framework to model the document as a graph with complex semantic relations rather than a simple linear chain.

Based on the constructed discourse graph, we apply a graph-driven method to select historical context for document-level translation dynamically.
Specifically, when translating one target paragraph, we perform a backward and biased random walk starting from the corresponding node. Then, we identify a related context path, which is composed of previous  paragraphs along with their translations.
Unlike previous approaches that restrict context to immediate neighbors \cite{dutta2025graft}, we guide the traversal with two complementary signals: local edge weights that encode semantic and lexical relevance, and a global signal that encourages traversing  nodes that can generate longer and richer discourse chains.
This design enables the model to select a single, structured context path that captures long-range, non-linear dependencies while remaining computationally efficient.
The selected path is subsequently formatted into a discourse-aware prompt, enabling the model to  exploit structured contextual information  without processing the entire text.
Moreover, since the traversal is probabilistic rather than deterministic, G$^{2}$C-MT naturally supports multi-path sampling. 
By exploring multiple plausible discourse paths and aggregating the resulting translations, the framework can improve robustness in the presence of discourse-level ambiguity

Our main contributions are as follows:
\begin{itemize}
\item We propose G$^{2}$C-MT, a novel framework that models document context as a  DAG,
which can be built in a lightweight way. The graph can capture non-linear discourse dependencies effectively  by modeling  semantic and sequential correlations.
\item  We design a biased random walk mechanism that explicitly favors deeper context paths containing  richer discourse information. Meanwhile, this probabilistic traversal 
can inherently support multi-path sampling, which can enhance robustness when the target paragraph involves  discourse  ambiguities.
\item  We conduct extensive experiments on various document-level translation benchmarks  using LLMs of different scales. Our method outperforms strong baselines in both translation quality and coherence. Further analysis confirms  the effectiveness of our graph-guided approach in capturing  long-range dependencies.
\end{itemize}

\section{Methodology}
As illustrated in Figure \ref{fig:overview} and Algorithm \ref{alg:g2c_mt}, the overall pipeline of our method can be divided into the following three stages:
\begin{enumerate}
\item Directed Discourse Graph Construction, which treats each paragraph as a  node and builds edges between paragraphs considering their  multi-dimensional relevance.
\item Depth-Biased Context Sampling, a stochastic process to backtrack previously translated paragraphs and favor deeper context paths.
\item Path-Based Contextual Generation, which formats these context paths as the discourse information to prompt LLMs.
\end{enumerate}

\subsection{Directed Discourse Graph Construction} 
We  model the source document $D = \{x_1, x_2, \dots, x_N\}$ as a weighted directed acyclic graph $G = (V, E)$ firstly.
The directed edge $e_{ij}$ connects a target node $v_i$ to a previous node $v_j$ where $j < i$. 
Note that  future translation $y_{i}$ (where $i > j$)  are unavailable at step $j$, which is consistent  with a human translating the document sentence by sentence. 
This definition of directed edges can also avoid the appearance of cyclic graphs, thus reducing complexity during traversing.

The weight $w_{ij}$ of edge $e_{ij}$ quantifies the relevance of paragraph $x_j$ to $x_i$, calculated by a fusion of three discourse-related factors:
\begin{equation}
w_{ij} = \alpha \cdot S_{sem}(i, j) + \beta \cdot S_{seq}(i, j) + \gamma \cdot S_{key}(i, j),
\end{equation}
where  $\alpha$, $\beta$, and $\gamma$ are coefficients, that sum to 1 to balance each factor.
The specific definitions are as follows:

\paragraph{Semantic Relevance ($S_{sem}$).}
Global coherence relies on thematic consistency. 
We map each paragraph $x_i$ into a dense vector space via a pre-trained embedding model, denoted as $\mathbf{h}_i$. Then we compute the cosine similarity between these vectors.
To prevent the graph from being too dense and introducing extra noise, we introduce a threshold $\tau_{\text{sem}}$  to truncate those edges with low correlation:

\begin{equation}
S_{sem}(i, j) = \max (0, \frac{\mathbf{h}_i^\top \mathbf{h}_j}{|\mathbf{h}_i| |\mathbf{h}_j|} - \tau_{\text{sem}})
\end{equation}

\paragraph{Sequential Adjacency ($S_{seq}$).}
The paragraph being translated is most closely related to its adjacent paragraphs. For example, in dialogue questionnaires or background introductions, adjacent contextual information is essential for reference resolution and for preserving logical coherence. Therefore, an adjacent edge is naturally introduced and assigned  a fixed weight:
\begin{equation}
S_{seq}(i, j) = \mathbb{I}(j = i-1)
\end{equation}
where $\mathbb{I}(\cdot)$ is the indicator function. This guarantees that the local context is always considered a candidate.

\paragraph{Keyword Overlap ($S_{key}$).}
Semantic-based retrieval may miss some paragraphs that contain overlapping keywords but have low semantic similarity. These omissions may lead to inconsistency in the translation of terms, such as some proper nouns. 
We alleviate this problem by introducing $S_{key}(i, j)$ of keyword overlap. Specifically, $\mathcal{K}_i$ denotes the set of top-$K$ keywords in $x_i$ extracted via TF-IDF. 
Then we calculate the lexical score through the degree of keyword overlap:
\begin{equation}
S_{key}(i, j) = \sum_{t \in \mathcal{K}_{ij}} \frac{\psi(t, x_i) + \psi(t, x_j)}{2}
\end{equation}
where $\mathcal{K}_{ij} = \mathcal{K}_i \cap \mathcal{K}_j$ denotes the intersection of keywords, and $\psi(t, x)$ represents the TF-IDF score of term $t$ in paragraph $x$.

\subsection{Depth-Biased Context Path Sampling} 
Once the discourse graph is built, we can backtrack the context path, starting from 
the given target paragraph $x_i$. The backtrack strategy is also important. The most straightforward method is greedy search, which always selects the neighbor node with the highest weight.  
However, we find that this method sometimes terminates prematurely, or tends to select some nodes with certain types of edges
,such as repeatedly traversing edges with highly similar semantics, resulting in redundancy.
To address this, we propose a sampling strategy to balance the edge relevance with the depth of context path  to exploit global structural information.

\subsubsection{Depth Heuristic}
We introduce a concept of \textit{Depth Heuristic} to estimate the informational richness of a context node. 
The Depth Heuristic $\phi(v_j)$  represents the longest path backtracked from node $v_j$.
We argue that this backtrace depth represents how much historical translation context can be provided from a given node $v_i$.
Specifically, we can efficiently calculate $\phi(v_j)$  via dynamic programming.

\begin{equation}
% \phi(v_j) = 1 + \max_{{v_k \mid (v_j, v_k) \in E_j} } \phi(v_k)
\phi(v_j) = 1 + \max_{v_k \in \mathcal{B}(v_j)} \phi(v_k)
\end{equation}
where $\phi(v_{start}) = 1$ and 
$\mathcal{B}(v_j)$ denotes the set of backward neighbors of $v_j$.

\subsubsection{Probabilistic Sampling}
We apply the random walk mechanism for context selection to construct the path $\mathcal{P}_i = (v_{p_1}, v_{p_2}, \dots, v_{p_L})$,  starting from the target $v_{p_1} = v_i$.
Given the current node $v_{curr}$, the walker transitions to a previous node $v_{next}$, which is sampled from its neighborhood $\mathcal{N}(v_{curr})$ defined above.
The transition probability is as follows:
\begin{equation}
P(v_{next} | v_{curr}) = \frac{w_{curr, next} \cdot (\phi(v_{next}) + 1)^\lambda}{Z}
\end{equation}
where $Z$ is the partition function for normalization, and the term $(\phi(v_{next}) + 1)^\lambda$ introduces a bias towards deeper structures.  
The hyperparameter $\lambda \geq 0$ affects the strength of the deep bias.  
When $\lambda = 0$, the traversal process degrades to a standard random walk based on edge weights. 
Increasing  $\lambda$ will favor nodes with higher depth, which encourages the retrieval of  long-range context.

\subsection{Path-Based Contextual Generation}
After completing the traversal,  we can employ this sampled discourse path $\mathcal{P}_i$ for in-context learning by prompting LLMs. 
We reverse the path to restore the natural document order, and the final prompt $I_i$ is constructed as follows:
\begin{equation}
I_i = \mathcal{I} \oplus [ (x_{p_L}, y_{p_L}) \oplus \dots \oplus (x_{p_2}, y_{p_2}) ] \oplus x_i
\end{equation}
where $\mathcal{I}$ denotes the translation instruction and $\oplus$ represents string concatenation. 
The pair $(x_k, y_k)$ denotes the previous source paragraph and its translation corresponding to the node $v_k$.

\paragraph{Multi-Path Sampling.}\label{multi_path}
Since our method is based on a random walk, each traversal can yield a different context path.
We can sample $K$ independent context paths $\{\mathcal{P}_i^{(1)}, \dots, \mathcal{P}_i^{(K)}\}$ for the target paragraph $x_i$
and then generate $K$ candidate translations $\{y_i^{(1)}, \dots, y_i^{(K)}\}$ accordingly.
The final translation can be determined via a majority voting mechanism or by selecting the candidate with the lowest perplexity. 
In this paper, we cluster the $K$ candidates via k-means and select the representative candidate closest to the cluster centroid,
which also proves to be a simple yet effective strategy.

\paragraph{Complexity Analysis.}
The time cost of graph construction primarily lies in embedding computation and TF-IDF keyword matching, leading to an overall time complexity of $O(N^2)$.
In practice, this is a one-time cost of $<$10\,s per typical document (e.g., $N\approx200$ sentences on a single CPU), and each random walk completes in milliseconds.
At inference time, G$^{2}$C-MT requires exactly $N$ LLM calls for $N$ paragraphs---the same as the Window-Based baseline---since graph construction and the walk are pre-processing steps that do not invoke the LLM.
Moreover, paragraphs below the similarity cutoff $\tau_{\text{sem}}$ are pruned, making the graph sparse.

\begin{algorithm}[t]
\caption{G$^{2}$C-MT: Graph-Guided Contextual Translation}
\small
\label{alg:g2c_mt}
\begin{algorithmic}[1]
\STATE \textbf{Input:} Source Document $D=\{x_1, \dots, x_N\}$, LLM $\mathcal{M}$
\STATE \textbf{Output:} Translated Document $Y$

\STATE \textbf{Stage 1: Directed Discourse Graph Construction}
\STATE Initialize $G=(V, E)$ with nodes $V=\{1, \dots, N\}$
\FOR{$i = 1$ \textbf{to} $N$}
    \FOR{$j = 1$ \textbf{to} $i-1$}
        \STATE Calc weight $w_{ij}$ via semantic/seq/keyword scores
        \IF{$w_{ij} > 0$}
            \STATE Add edge $(i, j)$ to $E$ with weight $w_{ij}$
        \ENDIF
    \ENDFOR
\ENDFOR

\STATE \textbf{Stage 2: Depth Heuristic Calculation}
\STATE Compute $\phi(v)$ for all $v \in V$ via dynamic programming

\STATE \textbf{Stage 3: Path-Based Generation}
\STATE $Y \leftarrow \emptyset$
\FOR{$i = 1$ \textbf{to} $N$}
    \STATE Sample path $\mathcal{P}_{\text{back}}$ from $x_i$ on $G$
    \STATE $\mathcal{P}_{\text{ctx}} \leftarrow \text{Reverse}(\mathcal{P}_{\text{back}} \setminus \{x_i\})$ 
    \STATE Construct prompt $I_i$ using $\mathcal{P}_{\text{ctx}}$ and $x_i$
    \STATE $y_i \leftarrow \mathcal{M}(I_i)$
    \STATE Append $y_i$ to $Y$
\ENDFOR
\STATE \textbf{Return} $Y$
\end{algorithmic}
\end{algorithm}

\section{Experiments}
\subsection{Experimental Setup}
\paragraph{Datasets}
We evaluate our models on two standard benchmark test sets for DocMT.
One of them is the SAP test set. Derived from the WAT 2020 and 2021 shared tasks, the set contains documents in the IT domain.
For this test set, experiments are conducted on six translation directions: English$\leftrightarrow$Vietnamese, English$\leftrightarrow$Chinese, and English$\leftrightarrow$Indonesian. The sentences are organized into separate documents, comprising approximately 2,000 sentences per direction.
We also employ the tst2017 test set from the IWSLT 2017 translation task, comprising parallel TED talk documents.
For this test set, we evaluate on the following eight translation directions: English$\leftrightarrow$Chinese, English$\leftrightarrow$French, 
English$\leftrightarrow$German, and English$\leftrightarrow$Japanese. 
Each translation direction consists of 10 to 12 sentence-aligned parallel documents, totaling approximately 1,500 sentences.
In both benchmarks, each document paragraph consists of a single sentence; thus each graph node corresponds to one sentence.

\paragraph{Baselines} We compare our model against the following baselines:
\begin{itemize}
\item \textbf{Sentence}: Each segment is translated independently  any contextual information.
\item \textbf{Window-Based}: A fixed-size sliding window of preceding sentences is used as context for translating the current sentence.
\item \textbf{Semantic-Based}: Context paragraphs are selected based on embedding-based semantic similarity to the source paragraph.
\item \textbf{GRAFT} \cite{dutta2025graft}:  A multi-agent framework for DocMT. The framework first determines the most relevant context paragraph and then extracts key alignment information, such as pronouns, entities, and phrases, to aid document-level translation.
\item \textbf{DelTA} \cite{wang2025delta}:  An agentic framework aimed at translation consistency, featuring a multi-level memory architecture responsible for proper nouns, summaries, and variable-term contexts. 
\end{itemize}
\paragraph{Settings}
For the backbone translation models, we utilize the APIs for Gemini-2.5-Flash-lite, DeepSeek-v3-0324, Qwen-2.5-72B-Instruct and Qwen3-235B-A22B-Instruct. To ensure reproducibility and deterministic outputs, we set the decoding temperature to $0$ for all LLMs.
For Graph Construction, we employ  \texttt{text-embedding-3-small} provided by OpenAI for the semantic edge calculation ($S_{sem}$). 
The semantic similarity threshold $\tau_{\text{sem}}$  is empirically set to $0.6$. The hyperparameters governing edge weight contribution are set to $\alpha=0.2$ (semantic), $\beta=0.3$ (sequential), and $\gamma=0.5$ (keyword), 
prioritizing lexical overlap and semantic coherence based on performance on the validation set. 
We verify term importance using TF-IDF, where each paragraph is treated as a document to compute the IDF within the scope of the source text.
For the Biased Random Walk, we set the path depth bias $\lambda = 2.0$ to encourage the selection of deeper context nodes. 
The maximum context path length (number of previous paragraphs included) is capped at $L=4$ and  stop the walk  when the distance exceeds 100.
For Window-Based and Semantic-Based baselines, we also set the context size to 4 paragraphs for a fair comparison.

% \paragraph{Evaluation} We report document-level BLEU (d-BLEU) scores with the standard \texttt{sacreBLEU} library.
% We also explore other discourse evaluation metrics  in the ablation studies. 

\begin{table*}[htbp]
\centering
\small
\sisetup{table-format=2.2}
\begin{tabular}{@{}llcccccc@{}} % @{} removes extra space at the edges
\toprule
\textbf{Model} & \textbf{Method} & \multicolumn{6}{c}{\textbf{d-BLEU Score}} \\ % 2. Changed multicolumn span from 7 to 6
\cmidrule(l){3-8} % 3. Changed cmidrule range from 3-9 to 3-8
& & {EN $\to$ VI} & {EN$\to$ZH} & {EN $\to$ ID} & {VI $\to$ EN} & {VI $\to$ ZH} & {ID $\to$ EN} \\ % 4. Removed the "Average" header
\midrule

% 4. Removed the average value from each row
\textbf{Gemini-2.5-Flash-lite} &Sentence & 65.7 & 40.0 & 55.9 & 53.3 & 37.4 & 52.7 \\
& Window-Based   & 66.4 & 44.9 &57.4 & 56.0 & 43.1 & 56.6 \\
& Semantic-Based & 65.8 &39.9 &55.8 & 55.1 & 41.2 & 56.3 \\
& \textbf{G$^{2}$C-MT}  &\textbf{67.3} & \textbf{45.1} & \textbf{58.2} & \textbf{57.2} & \textbf{43.8} & \textbf{57.2} \\
\midrule

\textbf{DeepSeek-v3-0324} & Sentence & 63.7 & 36.3 & 55.0 & 51.4 & 36.9 & 51.4 \\
& Window-Based   & 64.9 & 38.1 & 56.2 & 56.1 & 40.9 & 55.7 \\
& Semantic-Based & 64.5 & 37.7 & 54.6 & 54.7 & 39.8 & 54.2 \\
& \textbf{G$^{2}$C-MT}  & \textbf{65.5} & \textbf{39.2} & \textbf{56.9} & \textbf{56.6} & \textbf{41.8} & \textbf{56.9} \\
\midrule

\textbf{Qwen3-235B-A22B-Instruct} & Sentence & 63.0 & 39.6 & 52.8 & 52.7 & 37.6 & 52.7 \\
& Window-Based   & 63.8 & 40.8 & 54.4 & 55.4 & 40.9 & \textbf{55.3} \\
& Semantic-Based & 63.2 & 40.9 & 54.3 & 53.3 & 39.8 & 54.9 \\
& \textbf{G$^{2}$C-MT}  & \textbf{64.3} & \textbf{42.5} & \textbf{55.0} & \textbf{55.8} & \textbf{41.7} & 54.6 \\
\bottomrule
\end{tabular}
\caption{d-BLEU scores on the SAP benchmark across three different LLM backbones. Bold indicates the best performance. Results averaged over 3 random walk seeds ($\sigma \leq 0.3$). Improvements of G$^{2}$C-MT over Window-Based are significant ($p < 0.05$, bootstrap).}
\label{tab:sap_results}
\end{table*}

\begin{table*}[t]
\centering
\small
\begin{tabular}{@{} l *{8}{c} @{}} 
\toprule
\textbf{Method} & \multicolumn{8}{c}{\textbf{d-BLEU Score}} \\
\cmidrule(l){2-9} 
&{EN$\to$ZH}&{EN$\to$FR}&{EN$\to$DE}&{EN$\to$ JP}&{ZH$\to$EN}&{FR$\to$EN}&{DE $\to$ EN} & {JP $\to$ EN} \\ 
\midrule
NLLB-3.3B    & 30.8 & 34.7 & 24.3 & 13.7 & 26.4 & 42.8 & 33.1 & 16.6 \\
Google Translate    & 30.5 & 40.8 & 22.3 & 16.2 & 27.5 & 27.4 & 22.7 & 21.0 \\
\midrule
% \textcolor{red}{Doc2Doc}       & 35.8 & 40.5 & 27.5 & 17.0 & 27.9 & 42.0 & 32.1 & 18.5 \\
Sentence      & 36.5 & 41.4 & 28.0 & 17.6 & 28.3 & 42.4 & 32.8 & 19.3 \\
Window-Based       & \textbf{36.9} & 42.0 & 28.7 & 17.6 & 29.4 & 42.7 & 33.8 & 20.4 \\
Semantic-Based  & 36.6 & 41.7 & 28.4 & 17.8 & 29.1 & 43.1 & 34.3 & 17.9 \\
DelTA \cite{wang2025delta}        & 35.6 & 41.1 & 28.9 & 16.5 & 29.8 & \textbf{43.6} & 33.9 & 20.5 \\
GRAFT \cite{dutta2025graft}        & 36.3 & \textbf{44.8} & 28.5 & 16.1 & 30.1 & 43.1 & 34.3 & 17.9 \\ \hline
\textbf{G$^{2}$C-MT} & \textbf{36.9} & 42.0 & \textbf{29.1} & \textbf{18.1} & \textbf{30.6} & 43.4 & \textbf{34.5} & \textbf{21.0} \\
\bottomrule
\end{tabular}
\caption{d-BLEU scores on the IWSLT 2017 test set based on Qwen-2.5-72B-Instruct for fair comparison with prior work.}
\label{tab:iwslt_results}
\end{table*}

\subsection{Main Results}
\subsubsection{Performance on Technical Documentation}
The evaluation results of all translation directions and backbones on SAP datasets are presented in Table \ref{tab:sap_results}.
Firstly, we observe that the discourse context plays an important role in enhancing translation quality for DocMT.
The \textbf{Sentence} baseline consistently underperformes all context-aware methods, trailing the simple \textbf{Window-Based} context by an average of 2–3 d-BLEU.
Secondly, G$^{2}$C-MT consistently achieves the highest d-BLEU scores across all translation directions and backbones, 
validating the robustness and generality of our graph-based approach regardless of the underlying model architecture.
Thirdly, we observe that the \textbf{Window-Based} approach usually outperforms the \textbf{Semantic-Based} approach in this domain. 
This is intuitive for technical documentation, where logical progression (e.g., step 1, step 2) often implies that the adjacent sentences is the most relevant context. 
However, G$^{2}$C-MT surpasses the \textbf{Window-Based} baseline by substantial margins—for instance, achieving a gain of \textbf{+0.9 d-BLEU} on EN$\to$VI and \textbf{+1.2 d-BLEU} on VI$\to$EN using Gemini-2.5-Flash-lite. 
This indicates that even in highly sequential documents, our \textit{Depth-Biased Sampling} successfully retrieves necessary long-range constraints (such as terminology defined earlier in the document) that a fixed window misses, 
without introducing the noise associated with pure semantic retrieval.
We also note the unstable performance of the \textbf{Semantic-Based} method, which sometimes even underperforms the Sentence baseline (EN $\to$ ID).
This suggests that treating context as a set of independent fragments may have the negative effect of disrupting the logical flow, leading to incoherent translations.

\subsubsection{Performance on Narrative Discourse}
Table \ref{tab:iwslt_results} shows the results on the IWSLT 2017 benchmark using Qwen-2.5-72B-Instruct. 
This dataset is challenging for its loose conversational structure and long-range thematic dependencies.
First, G$^{2}$C-MT significantly outperforms the commercial and supervised baselines (NLLB-3.3B and Google Translate) across all directions, confirming the efficacy of LLMs for document-level translation when prompted with appropriate context.
Second, our method proves superior to recent state-of-the-art agentic frameworks,
which achieves comparable or higher d-BLEU scores (e.g., \textbf{+1.4} over GRAFT in EN$\to$ZH and \textbf{+2.1} over DelTA in JP$\to$EN). 
This result is significant given the computational efficiency of our method, 
as GRAFT and DelTA rely on computationally expensive, multi-step LLM calls to curate context or maintain memory modules.

\paragraph{Revisiting Context: Structure vs. Similarity.}
A notable pattern across both datasets is that the \textbf{Semantic-Based} baseline frequently fails to outperform the simple \textbf{Window-Based} approach, 
particularly on SAP (Table \ref{tab:sap_results}) and several IWSLT directions (e.g., EN$\to$DE, JP$\to$EN).
This highlights a critical fact in document translation: local cohesion often outweighs global topical relevance.
Pure semantic retrieval may break the linear narrative required for immediate syntactic dependency handling (e.g., pronoun resolution).
G$^{2}$C-MT avoids this trade-off by rooting retrieval in the discourse structure.
By explicitly modeling sequential edges ($S_{seq}$) alongside semantic ones, 
our graph traversal prioritizes the immediate history while the depth-biased sampling  allows the model to trace logical threads back to earlier context.

\section{Analysis}
In this section, we conduct a deeper analysis to investigate: 
(1) whether G$^{2}$C-MT effectively handles discourse phenomena; 
(2) how the model exploits long-range dependencies; 
(3) the potential of multi-path sampling for robustness; 
and (4) the contribution of each component through ablation studies.

\begin{table}[htbp]
\centering
\small
\begin{tabular}{@{}lcccc@{}}
\toprule
\textbf{Model}  & \textbf{BlonDe} & \textbf{d-Prism} & \textbf{d-Comet}& \textbf{d-BLEU}  \\
\midrule
Sentence & 44.77 & -1.97 & 2.29 & 47.8\\
Window-Based & 50.74 & -2.00 & 2.35 & 51.9 \\
Semantic-Based & 49.87 & -1.97 & 2.36 & 50.9\\
\textbf{G$^{2}$C-MT} & \textbf{51.35} & \textbf{-1.88} & \textbf{2.43}& \textbf{52.7} \\
\bottomrule
\end{tabular}
\caption{Evaluation of discourse Metric on the SAP dataset (EN $\to$ XX). Higher is better for all metrics.}
\label{tab:blonde_analysis}
\end{table}

\subsection{Evaluation of Discourse Metric}
\label{sec:discourse_analysis}
We adopt three specialized metrics to further investigate the discourse translation performance: \textbf{BlonDe} \cite{jiang-etal-2022-blonde}, which  explicitly tracks discourse phenomena such as  entities, tenses, and pronouns; \textbf{d-Prism} \cite{thompson-post-2020-paraphrase,easy_doc_mt}, which measures semantic consistency using probability scores; and \textbf{d-Comet} \cite{rei-etal-2022-comet,easy_doc_mt}, which evaluates translation quality by considering the preceding context.

As shown in Table \ref{tab:blonde_analysis}, G$^{2}$C-MT achieves the best performance across all metrics. 
\textbf{Window-Based} achieves high BlonDe scores but obtains lower d-Prism scores than \textbf{Semantic-Based}. This indicates that the \textbf{Window-Based} method  can capture local connections well, 
while the \textbf{Semantic-Based} method is good at finding relevant topics in long-range context. 
G$^{2}$C-MT mitigates this trade-off by obtaining the highest BlonDe score (51.35) while maintaining strong semantic consistency through the combination of sequential and semantic edges.

\begin{figure}[htbp]
    \centering
    \includegraphics[width=\columnwidth]{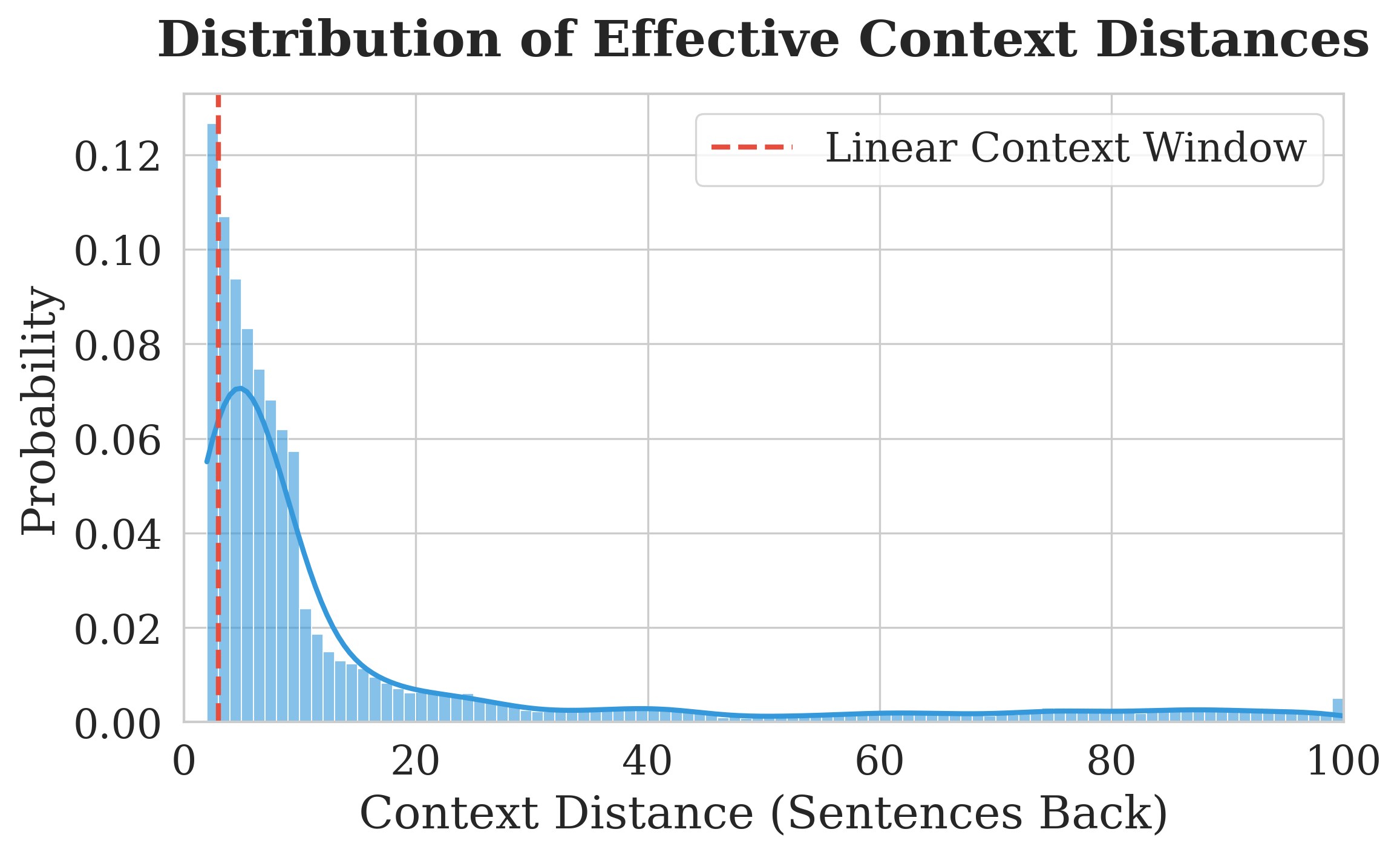}
    \caption{\textbf{Distribution of Context Distances.} 
    The red dashed line represents the hard cutoff of standard Window-based approach (Window=4). The blue histogram shows the context selected by G$^{2}$C-MT. Our method retains strong attention to local context (peak at $\Delta < 10$) while maintaining a long tail of retrieval capabilities extending up to 100 sentences back, capturing long-range dependencies that linear models miss.}
    \label{fig:context_dist}
\end{figure}

\subsection{Properites of Graph-Guided Context Selection}
\label{sec:context_analysis}
To better understand the efficacy of the proposed pipeline, 
we characterize the context selection behavior of G$^{2}$C-MT compared to the \textbf{Semantic-Based} and \textbf{Window-Based} methods. 
We define the \textit{Context Distance} $\Delta = i - k$ and visualize the distribution of the selected nodes in Figure \ref{fig:context_dist}.
We can observe two key properties of G$^{2}$C-MT from the figure:

\begin{itemize}
\item \textbf{Beyond Fixed Windows:}
The \textbf{Window-Based} method uses a fixed window size for context selection, where any dependency beyond the window size $L$ is inaccessible.
Our method G$^{2}$C-MT can walk through semantic edges ($S_{sem}$) to skip over intermediate nodes. 
We can see that in Figure \ref{fig:context_dist}, G$^{2}$C-MT effectively captures context at distances $\Delta > 20$.
\item \textbf{Beyond Isolated Retrieval (Coherence over Similarity):} 
the \textbf{Semantic-Based} method retrieves context based on similarity scores, resulting in a \textit{bag of sentences} $\{x_{k}\}$ without structural connections. 
However, the contexts selected by G$^{2}$C-MT are the components of  a \textit{connected path} $\mathcal{P}_i$. 
There exists a strong connection between nodes, since these edges of the path are weighted by discourse factors (Semantic, Sequential, and Lexical).
The explicitly modeled sequential edges ($S_{seq}$) act as glue, so the distribution of context distances in G$^{2}$C-MT still peaks at small $\Delta$ values.
\end{itemize}

\begin{table}[t]
\centering
\small
\begin{tabular}{@{}lcc@{}}
\toprule
\textbf{Strategy} & \textbf{d-BLEU}  \\
\midrule
Single Path ($K=1$) & 67.3   \\
Multi-Path ($K=3$)  & 67.8   \\
Multi-Path ($K=5$)  & \textbf{67.9}   \\
\bottomrule
\end{tabular}
\caption{Effect of Multi-Path Exploration on translation quality (En$\to$Vi).}
\label{tab:multipath}
\end{table}
\subsection{Multi-Path Sampling for Robustness}
In this section, we analyze the effect of multi-path sampling in G$^{2}$C-MT.
As introduced in Section \ref{multi_path}, 
we first sample multiple context paths for the same target paragraph by random walks,
and then select the final translation from multiple transslation candidates.
We use the embedding model to vectorize each candidate translation, and then select the candidate closest to the cluster centroid as the final output.
We experimented with sampling $K=\{1, 3, 5\}$ paths on the difficult En$\to$Vi subset, leaving larger $K$ values for future work due to inference cost.
As shown in Table \ref{tab:multipath}, 
there is a consistent improvements in d-BLEU as we increase the number of sampled paths $K$.
When $K=5$, we achieve the best performance of 67.9 d-BLEU, which is +0.6 higher than the single-path baseline. 
This suggests that when translating ambiguous sentences, different context paths may provide complementary information, leading to more robust translations.
Although this comes at the cost of increased inference latency, it offers a flexible trade-off for scenarios where quality is preferable.

\begin{table}[htbp]
\centering
\begin{tabular}{@{}lcc@{}}
\toprule
\textbf{Method} & \textbf{En $\to$ XX } & \textbf{$\Delta$} \\
\midrule
\textbf{G$^{2}$C-MT (Full)} & \textbf{56.9} & - \\
\midrule
\quad w/o Keyword Edge ($S_{key}$)   & 56.4 & -0.5 \\
\quad w/o Semantic Edge ($S_{sem}$) & 56.7 & -0.2 \\
\quad w/o Sequential Edge ($S_{seq}$)& 56.5 & -0.4 \\
\quad w/o Biased Random Walk    & 56.4 & -0.5 \\
\bottomrule
\end{tabular}
\caption{Ablation study on edge components and search strategy on Gemini-2.5-Flash-lite.}
\label{tab:ablation}
\end{table}

\subsection{Ablation Study}
We conduct an ablation study on the SAP dataset to analyze the effect of different edge types and our proposed search strategies. The results are summarized in Table \ref{tab:ablation}. 
From the table, we can observe that removing the \textbf{Keyword Edge} or the \textbf{Biased Random Walk} results in the largest performance drops (-0.5).
This result is intuitive, as terminological consistency is crucial in technical documents, especially for the SAP dataset which is in the IT domain.
The \textbf{Biased Random Walk} also plays an important role in preventing the model from settling for shallow, uninformative context.
Another  noticeable observation  is that the absence of the \textbf{Sequential Edge} ($S_{seq}$) causes a larger performance drop (-0.4) compared to removing the \textbf{Semantic Edge} ($S_{sem}$) (-0.2).
This indicates that local sequential order is crucial for immediate discourse coherence (e.g., pronoun resolution), while  the less impact of semantic edges may imply that the \textbf{Keyword Edge} has already captured much of the necessary topical relevance.

\subsection{Case Study}
Table \ref{tab:case_study} demonstrates how G$^{2}$C-MT handles long-range ambiguity.
The source term \textit{posting} is polysemous: it  means publishing but refers to accounting entry in this specific ERP context.
The defining context appears ten paragraphs earlier, causing the window-based baseline to default to the generic, incorrect translation.
In contrast, G$^{2}$C-MT successfully retrieves the prior paragraph guided by keyword overlap edges—and generates the correct domain terminology.

\begin{table}[htbp]
\centering
\small
% \caption{A case study on long-range lexical disambiguation in SAP dataset. The standard window-based baseline misinterprets the polysemous term due to lack of context, whereas G$^{2}$C-MT successfully retrieves the defining paragraph from distance -10 to resolve the ambiguity.}

\begin{CJK*}{UTF8}{gbsn}
\resizebox{\columnwidth}{!}{
\begin{tabular}{l p{7.5cm}}
\toprule
\multicolumn{2}{c}{\textbf{Contextual Information}} \\
\midrule
\textbf{Distant Context} & \textit{Scenario: Automatic \textbf{\textcolor{teal}{Posting}}} \\
(Dist: -10 paras) & Represents a business context where... processes can trigger automatic \textbf{\textcolor{teal}{postings}}, for example, elimination \textbf{\textcolor{teal}{posting}}, adjustment \textbf{\textcolor{teal}{posting}}... \\
\textbf{Source Input ($x_i$)} & If \textbf{jobs} have been scheduled for \textbf{posting} periods, you can change them by choosing this button. \\
\midrule
\multicolumn{2}{c}{\textbf{Translations}} \\
\midrule
\textbf{Reference} & 如果已为\textbf{过账}期间计划\textbf{作业}，则可通过选择此按钮进行更改。 \\ \hline
\textbf{Baseline} & 如果已为\textbf{\textcolor{red}{发布}}期间安排了\textbf{\textcolor{red}{任务}}，则可以通过选择此按钮来更改它们 。 \\
\textit{(Window-based)} & \textit{\footnotesize{(Fails to access distinct context, treating "posting" as "publish")}} \\ 
\hline
\textbf{G$^{2}$C-MT} & 如果已为\textbf{\textcolor{teal}{过账}}期间安排了\textbf{\textcolor{teal}{作业}}，则可以通过选择此按钮来更改它们。 \\
\textit{(Ours)} & \textit{\footnotesize{(Retrieves context via graph, correct domain terminology)}} \\
\bottomrule
\end{tabular}
}
\end{CJK*}
\caption{A case study on long-range lexical disambiguation in SAP dataset.}
\label{tab:case_study}
\end{table}

\section{Related Work}
\subsection{LLM-based Document-Level MT}
LLMs have demonstrated strong in-context learning and long-context modeling capabilities,
which makes them suitable for DocMT \cite{Wang2023DocumentLevelMT,wu2024adapting,cui2024efficiently}. 
One recent line of work focuses on combining graph structures with LLMs to generate discourse-aware translations 
since the graph can naturally model discourse dependencies. 
\citeauthor{dutta2025graft} proposed GRAFT, which uses an LLM agent to segment documents into discourse units, identify context dependencies, and form a directed acyclic discourse graph.
TransGraph \cite{pham2025discourse} similarly models inter-chunk discourse relations via LLM-based relation classification.
Our work differs from both along three axes:
\textbf{(i)~Graph construction cost}---GRAFT and TransGraph require $O(N^2)$ LLM calls for pairwise edge classification, whereas G$^{2}$C-MT constructs edges using lightweight embedding similarity and TF-IDF in a single pass;
\textbf{(ii)~Context depth}---prior methods restrict context selection to first-order neighbors, while our depth-biased random walk discovers multi-hop paths spanning up to 100 nodes;
\textbf{(iii)~Multi-path robustness}---stochastic traversal enables multi-path sampling (Table~\ref{tab:multipath}), a capability absent in prior graph-based DocMT.
Another line of work focuses on building memory mechanisms \cite{wang2025delta} to maintain document-level consistency during translation, but these approaches tend to underemphasize explicit discourse structure modeling.

\subsection{Structured Reasoning and Self-Consistency in LLMs}
LLMs' reasoning capabilities can be effectively invoked through sophisticated prompting techniques instead of simple instruction-response paradigms.
Chain-of-Thought (CoT) prompting \cite{wei2022chain} shows the potential of handling complex reasoning tasks by generating intermediate reasoning steps.
Building on this work, non-linear reasoning frameworks have been proposed, such as Tree of Thoughts (ToT) \cite{yao2023tree} and Graph of Thoughts (GoT) \cite{besta2024graph},
further improving reasoning performance by exploring multiple reasoning paths.
Beyond structural exploration, single-path generation can be extended to multi-path generation to enhance robustness via self-consistency \cite{wang2022self}.
This paradigm is particularly suitable for DocMT,  since the translation of a given paragraph often relies on ambiguous discourse clues that may permit multiple valid interpretations.
In this work, we bridge these mechanisms by integrating graph-based structural modeling with multi-path  sampling for DocMT.

\subsection{Retrieval-based Machine Translation}
Early works in NMT have explored retrieval-based methods to improve domain adaptation and translation quality.
One representative line of work \cite{khandelwal2021nearest,meng-etal-2022-fast} retrieves token-level examples from a vector datastore to calibrate the model's output distribution.
With the powerful in-context learning ability of LLMs, retrieving few-shot examples in sentence-level for LLM has become a dominant paradigm \cite{agrawal2023context,ji2024submodular,zebaze2025context}.
\cite{wang2025delta,cui2024efficiently} work on document-level translation by retrieving relevant examples but still lack explicit modeling of discourse structure and primarily consider semantic similarity,
treating the document as a bag of unrelated sentences.

\section{Conclusion}
In this paper, we presented G$^{2}$C-MT, a novel framework for document-level machine translation.
Our method models the document as a directed graph to capture structured discourse dependencies.
By using depth-biased random walks, we select high-quality context paths as discourse context for prompting LLMs.
Experiments across technical and narrative domains show that G$^{2}$C-MT significantly outperforms strong baselines.
Furthermore, our approach is computationally efficient compared to complex agent-based methods.
We believe this graph-guided strategy provides a robust solution for long-text translation tasks.

\section*{Acknowledgments}
We would like to thank the anonymous reviewers for the helpful comments. This work was 
supported by National Natural Science Foundation of China (Grant No. 62276179, 62537001) and 
Project Funded by the Priority Academic Program Development of Jiangsu Higher EducationInstitutions.
Xiangyu Duan is the corresponding author.
\bibliographystyle{named}
\bibliography{ijcai26}

\end{document}